\documentclass[letterpaper, 10 pt, conference]{ieeeconf}  % Comment this line out if you need a4paper

\IEEEoverridecommandlockouts                              % This command is only needed if 
\usepackage[bookmarks=true]{hyperref}
\usepackage{tabularx}
\usepackage{graphicx}
\usepackage{epsfig}
\usepackage{subfig}
\usepackage{cite}
\usepackage{float}
\usepackage{ifthen}
\usepackage{xcolor}
\usepackage{ntheorem}
\usepackage{xcolor}
\theoremseparator{:}
\usepackage{soul}
\usepackage{colortbl}
\usepackage{booktabs}
\usepackage{multirow}
\usepackage{amsfonts}
\usepackage[normalem]{ulem}  % Allows strikethrough

\usepackage[font=footnotesize]{caption}
\usepackage[flushleft]{threeparttable}
\usepackage{enumerate}
\usepackage[acronym]{glossaries}
\newacronym{rmis}{RMIS}{robot-assisted minimally invasive surgery}
\newacronym{urdf}{URDF}{Unified Robot Description Format}
\newacronym{dvrksi}{dVRK Si}{da Vinci Research Kit Si}
\newacronym{ros}{ROS}{the Robot Operating System}
\newacronym{psm}{PSM}{patient side manipulator}
\newacronym{dof}{DoF}{degrees of freedom}
\newacronym{ecm}{ECM}{endoscopic camera manipulator}

\newcommand{\revise}[1]{{#1}}
\newcommand{\deleted}[1]{}

\title{\LARGE \bf
SurgiPose: Estimating Surgical Tool Kinematics from Monocular Video for Surgical Robot Learning
}

\author{Juo-Tung Chen$^{1}$, XinHao Chen$^{1}$, Ji Woong Kim$^{1}$, Paul Maria Scheikl$^{1}$, \\
Richard Jaepyeong Cha$^{2}$, and Axel Krieger$^{1}$
\thanks{$^{1}$ Dept. of Mechanical Engineering, Johns Hopkins University, Baltimore, MD, USA {\tt\small \{jchen396, xchen254, jkim447, pscheik1, axel\}@jhu.edu}}%
\thanks{$^{2}$Optosurgical, Columbia, 21046, USA {\tt\small JCHA2@childrensnational.org}}}%

\begin{document}

\maketitle
\thispagestyle{empty}
\pagestyle{empty}

%%%%%%%%%%%%%%%%%%%%%%%%%%%%%%%%%%%%%%%%%%%%%%%%%%%%%%%%%%%%%%%%%%%%%%%%%%%%%%%%
\begin{abstract}

Imitation learning (IL) has shown immense promise in enabling autonomous dexterous manipulations, including in learning surgical tasks. To fully unlock the potential of IL for surgery, access to clinical datasets is needed, which unfortunately lack the kinematic data required for current IL approaches.
A promising source of large-scale surgical demonstrations is monocular surgical videos available online, making monocular pose estimation a crucial step toward enabling large-scale robot learning. Towards this end, we propose SurgiPose, a differentiable rendering-based approach to estimate kinematic information from monocular surgical videos, eliminating the need for direct access to ground-truth kinematics. Our method infers tool trajectories and joint angles by optimizing tool pose parameters to minimize the discrepancy between rendered and real images. To evaluate the effectiveness of our approach, we conduct experiments on two robotic surgical tasks—tissue lifting and needle pickup—using the da Vinci Research Kit Si (dVRK Si). We train imitation learning policies with both ground-truth measured kinematics and with estimated kinematics from video and compare their performance. Our results show that policies trained on estimated kinematics achieve comparable success rates to those trained on ground-truth data, demonstrating the feasibility of using monocular video-based kinematic estimation for surgical robot learning.
By enabling kinematic estimation from monocular surgical videos, our work lays the foundation for large-scale learning of autonomous surgical policies from online surgical data.

\end{abstract}

%%%%%%%%%%%%%%%%%%%%%%%%%%%%%%%%%%%%%%%%%%%%%%%%%%%%%%%%%%%%%%%%%%%%%%%%%%%%%%%%

\section{Introduction}

Estimating the precise 6 Degrees of Freedom (DoF) pose of articulated surgical instruments from endoscopic images is a fundamental challenge in \gls{rmis}. Accurate pose estimation is crucial for surgical skill assessment~\cite{lajko2021endoscopic,elek2022towards} and workflow analysis~\cite{zia2018surgical,kitaguchi2020real}, as it provides insights into instrument motion patterns and procedural efficiency. \deleted{Additionally, }Pose estimation \revise{also} plays a key role in both model-based~\cite{joglekar2025autonomous,hari2024stitch,hu2024tissue,afshar2022tissuemanipulation} and learning-based~\cite{haiderbhai2024sim2real,tanwani2021sequential,pore2021learning,scheikl2024mpd} approaches for autonomous surgery. In particular, imitation learning can greatly benefit from accurate pose estimation, as it relies on precise motion data to map visual inputs to \deleted{motor }actions.
Additionally, with the growing interest in large-scale vision-language-action (VLA) models~\cite{kim24openvla,black2024pi_0}, obtaining expert demonstration data for imitation learning at scale has become increasingly crucial. Toward this end, one promising strategy is to extract kinematics data from robotic surgery videos that are widely available on the web \cite{schmidgall2024generalsurgeryvisiontransformer}. These videos often show monocular footage rather than stereo, since they are intended for demonstration or educational purposes. This motivates the development of alternative approaches that can infer accurate instrument motion solely from monocular video data, enabling scalable learning from real surgical demonstrations.

\begin{figure}
    \centering
    \vbox{
    \includegraphics[width=1\linewidth]{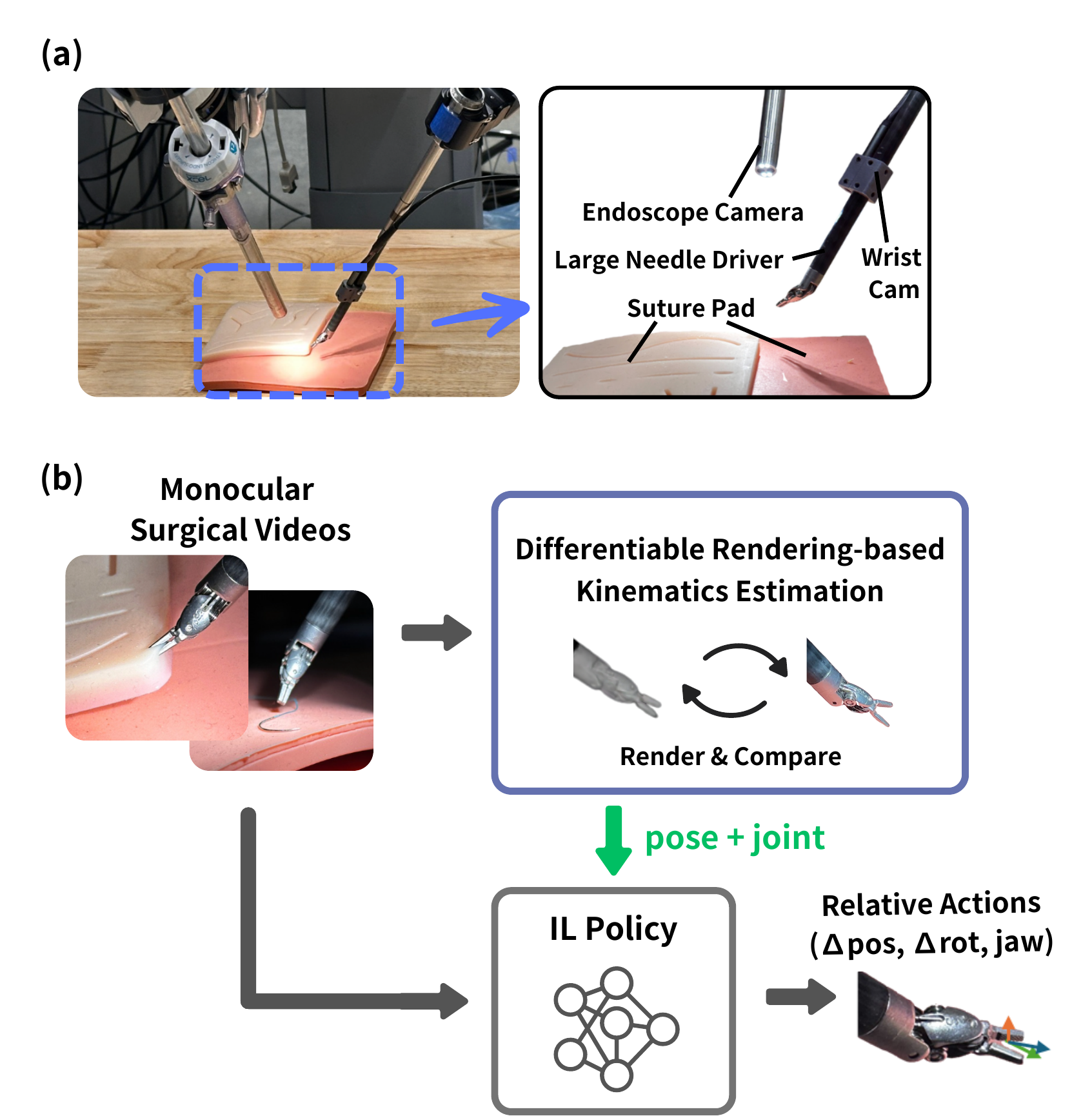}
    }
    \caption{(a) System setup (b) Overall workflow of our approach. Monocular surgical videos are processed by SurgiPose to infer kinematic information (tool poses and joint angles). The estimated kinematics, along with video frames, can then be used to train imitation learning policies, outputing actions for autonomous execution of surgical tasks.}
    
    \label{fig:overview}
\end{figure}

\begin{figure*}[t]
    \centering
    \vbox{
    \includegraphics[width=\textwidth]{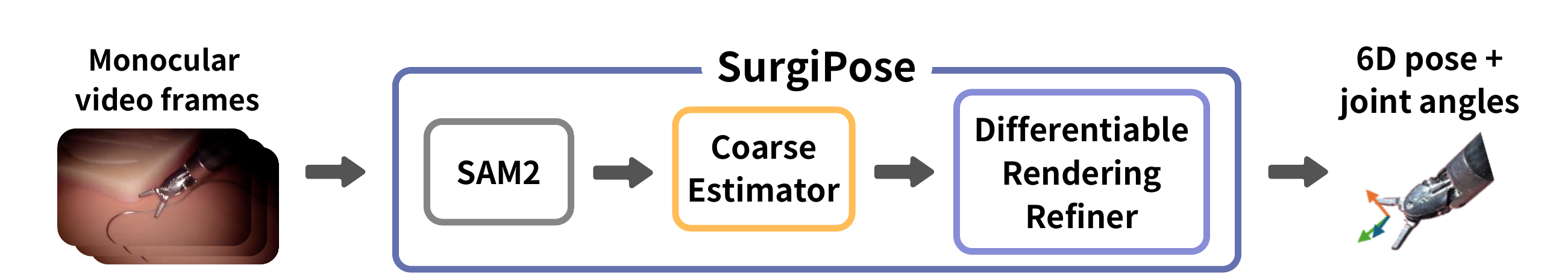}
    
    }
    \caption{Overview of the SurgiPose pipeline. The first stage (coarse estimator) initializes the pose, which is then refined via differentiable rendering. The final estimated 6-DoF pose and joint angles are used for kinematic extraction and imitation learning.}
    \label{fig:system_flowchart}
\end{figure*}

Despite the potential of trajectory estimation from video, accurately extracting tool motion remains challenging due to occlusions, lighting variations, and complex articulated motion in surgical environments. Traditional approaches rely on fiducial markers~\cite{zhao2016fiducial} or manually annotated keypoints~\cite{lu2021super,wang2020image} to estimate pose, but these methods are impractical in real surgeries due to setup constraints and potential interference with the procedure. Other techniques leverage stereo vision~\cite{hao2018vision}, depth sensing~\cite{ogor20193d}, or kinematic information~\cite{reiter2014appearance,ye2016real,hao2018vision,lu2021super}, yet these approaches require additional hardware or manual initialization~\cite{allan20183}, 
% limiting their applicability to unstructured surgical videos.
limiting their applicability with existing robotic surgery video datasets. Furthermore, these methods typically require more than just monocular images, such as stereo images or kinematics data, which is typically not available in the internet-scale videos online.
While learning-based tracking methods have shown promise, many focus only on non-articulated tools~\cite{hu2024weakly} or predict 2D keypoints instead of full 6 DoF poses~\cite{kong2023automatic,park2024towards}.
% , restricting their effectiveness for modeling complex articulated motion.
More recent methods~\cite{labbe2021single} focusing on articulated tools have explored render-and-compare strategies. For instance, differentiable rendering techniques~\cite{liudifferentiable} have demonstrated the effectiveness of Gaussian splatting for learning articulated robot models and reconstructing both 6 DoF pose and joint angles.
However, relying solely on differentiable rendering for pose estimation can be challenging, as poor initial pose estimates may lead to optimization failures and inaccurate reconstructions~\cite{tremblay2023diff}.

%% Our method
To address these challenges, we propose SurgiPose, a differentiable rendering pipeline for extracting surgical instrument trajectories from monocular videos, as shown in Fig.\ref{fig:overview}. Our method optimizes and estimates 6 DoF pose and joint angles leveraging differentiable rendering. The core idea behind differentiable rendering is that by making the rendering process continuous and differentiable, we can compute gradients that allow us to iteratively refine the estimated pose to better match the observed image. This render-and-compare optimization strategy enables marker-less, hardware-free motion extraction, making it well-suited for learning robot control policies from online surgical videos or expert demonstrations. This capability is crucial for scaling up imitation learning and building large-scale vision-language-action (VLA) models, as it allows kinematic data to be extracted from any publicly available robotic surgery video dataset. By eliminating the dependence on robot kinematics, our method broadens access to expert demonstrations, facilitating the development of data-driven autonomous surgical systems.

Our main contributions are: 

\begin{enumerate}
    \item A framework for leveraging monocular surgical videos to generate kinematic data at scale, reducing reliance on motion capture systems and enabling internet-scale surgical robot learning.
    \item A novel monocular 6 DoF pose estimation approach that combines coarse estimation with differentiable rendering, where the coarse estimator provides a crucial pose initialization to improve robustness and accuracy.
    \item Experiments demonstrating the feasibility of learning imitation policies from estimated kinematics, with performance comparable to ground-truth-based policies.
\end{enumerate}

\section{Materials And Methods}

%% Method overview
Our framework estimates the 6 DoF pose of a surgical tool's end-effector relative to the camera frame, denoted as $T_{CE} \in SE(3)$. We adopt a two-stage pose estimation approach, which is commonly used for pose estimation~\cite{labbe2022megapose,tremblay2023diff}. In our pipeline, the first stage generates an initial pose estimate using a coarse estimation module and then the second stage refines it through differentiable rendering.

The workflow of SurgiPose is illustrated in Fig.~\ref{fig:system_flowchart}. 

We first segment and crop surgical tools using SAM2~\cite{ravi2024sam2}. If the image is the first frame, we generate an initial pose estimate using a coarse estimation module. We then refine this estimate using differentiable rendering, optimizing both tool pose and joint angles to minimize discrepancies between rendered and observed images.
For video sequences, after processing the first frame with our coarse estimation and differentiable rendering modules, we use the refined pose as the initialization for subsequent frames. To ensure the refiner effectively tracks the pose in each frame, we perform up to 10 iterations per frame. If the loss plateaus, we apply early stopping and proceed to the next frame.

\subsection{Coarse Pose Estimation}
Differentiable rendering relies on an iterative optimization process to refine pose estimates by minimizing the difference between rendered and observed images. However, if the initial pose estimate is too far from the true pose, the optimization can become trapped in incorrect local minima, leading to failure. To address this, we introduce a coarse pose estimator that provides a robust initial guess, ensuring stable convergence and improving overall accuracy.
In this study, the term “initial guess” refers to the estimation of the tool pose in the first frame of a video. This step is critical because subsequent frames rely on tracking the pose from the previous frame. Therefore, an inaccurate initialization can propagate errors throughout the sequence, significantly degrading performance.

\begin{figure}[t]
    \centering
    \vbox{
        \includegraphics[width=1\linewidth]{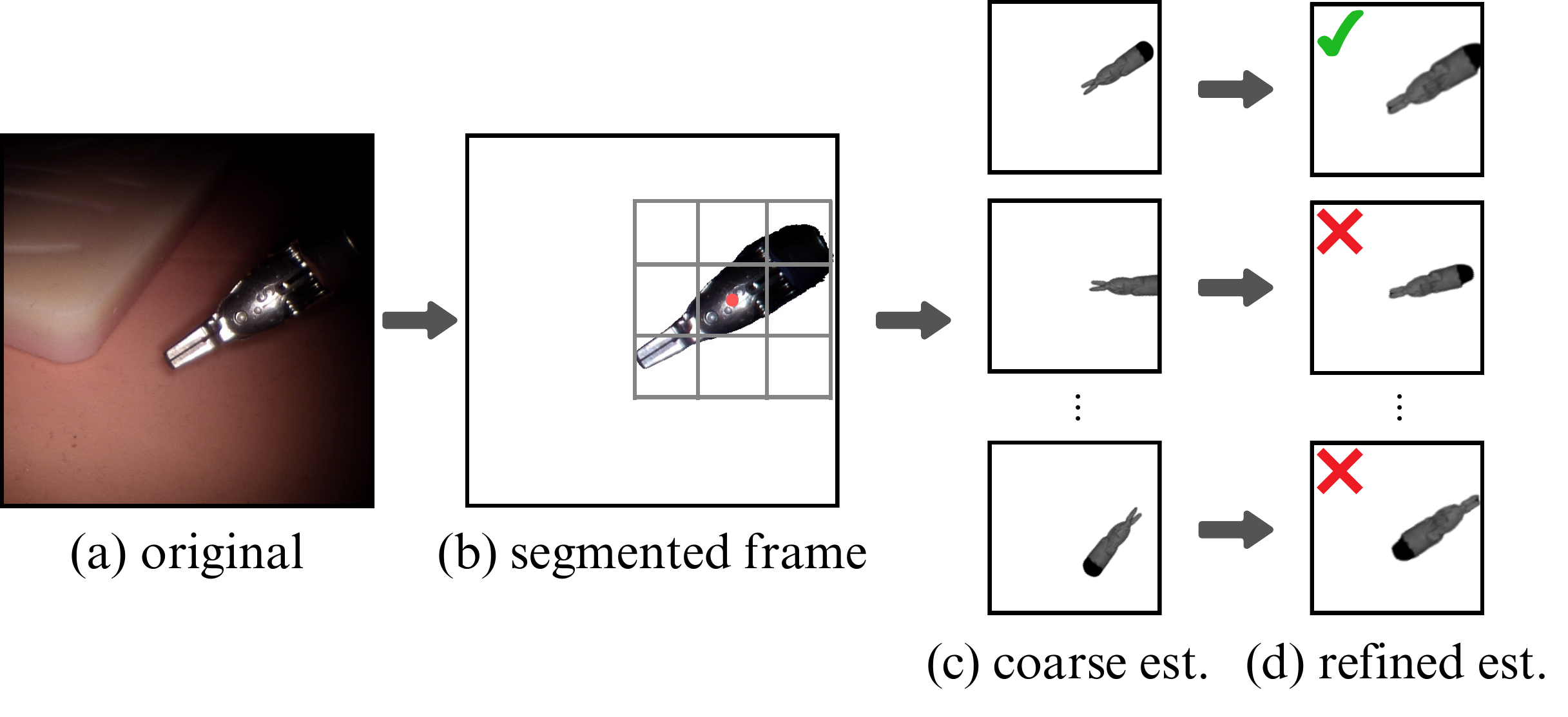}
    }
    \caption{Visualization of the coarse estimation pipeline. (a) Original video frame. (b) Segmented and cropped surgical tool with the calculated center of the mask and the corresponding 3x3 grid for proposing potential initial guesses. (c) Coarse estimations proposed by the coarse estimation module. (d) Selecting the best initial guess based on the lowest loss among the refined estimations.}
    \label{fig:estimation-stages}
\end{figure}

As illustrated in Fig.~\ref{fig:estimation-stages}, our coarse estimator generates multiple candidate initial poses and selects the best one based on rendering loss. The process is as follows: First, the center of the tool in the first frame is calculated based on the segmented tool mask. Using this center as the midpoint, a 3×3 square grid is constructed parallel to the image plane. Then, for each point on the grid, 36 potential initial guesses are generated by applying z-axis (pointing into the image) rotations with angles uniformly distributed between 0 and 2$\pi$. Note that rotations about the x (to the right) and y (downwards) axes and depth variations are not considered, as the subsequent refining process can resolve these parameters effectively. Each trial is refined via differentiable rendering, and the corresponding pixel-averaged loss is computed. This loss metric helps to eliminate bias toward guesses closer to the camera, which yield more pixels. The trial with the lowest loss is selected as the initial guess $T_{CE}^{initial}$.
By systematically evaluating multiple hypotheses, our coarse estimator ensures that the optimization starts from a pose close to the true tool pose, significantly reducing the risk of failure and improving downstream kinematics estimation.

%% --------- Training procedure ------------
\subsection{Differentiable Rendering}
To refine the initial guess, we train a differentiable model of the surgical tool using a synthetic dataset generated in MuJoCo. We put the URDF of the surgical tool\footnote{\url{https://github.com/jhu-dvrk/dvrk_model/tree/main}} into MuJoCo simulation, and we generate 500 canonical tool poses, where the joint angles remain fixed in neutral position, and 10,000 pose-conditioned tool configurations, where joint angles are randomly sampled from non-self-collision configuration. Each configuration is rendered from 12 random camera viewpoints with varying azimuth, elevation, and distance. The dataset includes the following: 

\begin{itemize}
\item Joint positions: Represented as \(\mathbf{q} = [q_1, q_2, q_3]^\top \in \mathbb{R}^3\), where \(q_1\) corresponds to the pitch of the tool, and \(q_2, q_3\) represent the angles of the two jaws. Since we focus only on the end-effector part of the CAD model, the tool has a total of 3 DoF in joint angles.

\item Camera extrinsic parameters: Given by a transformation matrix \(\mathbf{T}_c^w \in SE(3)\), which maps points from the world frame to the camera frame:
  \[
  \mathbf{T}_c^w = 
  \begin{bmatrix}
  \mathbf{R}_c^w & \mathbf{t}_c^w \\
  0 & 1
  \end{bmatrix}
  \]
  where \(\mathbf{R}_c^w \in SO(3)\) is the rotation matrix, and \(\mathbf{t}_c^w \in \mathbb{R}^3\) is the translation vector.
\item Camera intrinsic parameters: Modeled by the intrinsic matrix \(\mathbf{K} \in \mathbb{R}^{3 \times 3}\), which maps 3D points in the camera frame to 2D image coordinates:
  \[
  \mathbf{K} =
  \begin{bmatrix}
  f_x & 0 & c_x \\
  0 & f_y & c_y \\
  0 & 0 & 1
  \end{bmatrix}
  \]
  where \(f_x, f_y\) are the focal lengths, and \((c_x, c_y)\) represents the principal point of the camera.

\item Depth images converted into point clouds 
\item Rendered images
\end{itemize}

%% ----- Training stages --------
We train the differentiable model in three stages following \cite{liudifferentiable}. First, a canonical 3D Gaussian representation is learned to reconstruct a high-fidelity, static version of the tool from multi-view images. Next, a deformation field is introduced to model shape variations caused by different tool configurations. Finally, joint training optimizes both the canonical model and deformation field to improve accuracy under varying joint configurations.

The training process is implemented using PyTorch and leverages differentiable Gaussian splatting based on the open-source implementation from \cite{kerbl3Dgaussians}. We train the model on an NVIDIA RTX 4090 GPU, using the same hyperparameter settings as in \cite{liudifferentiable}. The model is trained for 20,000 iterations. Since the end effector is significantly smaller than the shaft, we modify the URDF model by shorten the shaft to better capture fine-grained details of the jaw’s appearance and motion during training.

\subsection{Optimization}
At test time, the refiner module takes $T_{CE}^{initial}$ and refines both  $T_{CE}^{initial}$ and joint angles using differentiable rendering. The optimization process updates the pose estimate by minimizing the difference between the rendered and observed images through gradient-based optimization.

The objective function combines structural similarity (SSIM) and mean squared error (MSE) to balance perceptual quality and pixel-wise accuracy:
\[
L_{\text{combined}} = \alpha (1 - \text{SSIM}(I_{\text{ren}}, I_{\text{obs}})) + (1 - \alpha) \| I_{\text{ren}} - I_{\text{obs}} \|_2^2
\]

where \( I_{\text{ren}} \) is the image synthesized by the differentiable renderer, \( I_{\text{obs}} \) is the real captured image, and \( \alpha \) controls the trade-off between SSIM loss and MSE loss. We set \( \alpha = 0.8\) empirically.

To ensure stable optimization, we use a learning rate scheduler that reduces the step size when the loss plateaus. The reduction factor was set to be 0.5 and patience set to be 20 epochs. Early stopping is applied if the loss change falls below a predefined threshold of \(1 \times 10^{-7}\) over 10 iterations. Additionally, to prevent extreme updates, translation gradients are clamped within \([-0.02, 0.02]\) range.

The pose parameters are updated iteratively using:

\[
\begin{aligned}
R_{\text{updated}} &= R_{\text{current}} \cdot (\mathbf{I} + \alpha \nabla R), \\
t_{\text{updated}} &= t_{\text{current}} - \beta \cdot \text{clamp}(\nabla t, -\delta, \delta)
\end{aligned}
\]

where:
\begin{itemize}
    \item \( R_{\text{updated}} \) is the updated rotation matrix after each optimization step.
    \item \( R_{\text{current}} \) is the rotation matrix before applying the update.
    \item \( \mathbf{I} \) is the \( 3 \times 3 \) identity matrix.
    \item \( \nabla R \) is the gradient of the loss function with respect to the rotation matrix.
    \item \( t_{\text{updated}} \) is the updated translation vector after each optimization step.
    \item \( t_{\text{current}} \) is the current translation vector before applying the update.
    \item \( \alpha \) is the learning rate for updating the rotation matrix.
    \item \( \beta \) is the learning rate for updating the translation vector.
    \item \( \delta \) is the clamping threshold to restrict extreme translation updates.
\end{itemize}

We set \( \alpha = 0.3\), \( \beta = 3 \times 10^{-4}\), and \( \delta = -0.02\) through experiment and fine-tuning.
The transformation matrix is enforced to remain a valid homogeneous transformation after each update, ensuring numerical stability and preventing divergence during optimization.

In addition to updating the end-effector pose, we optimize the joint angles \(\mathbf{q}\) to minimize the rendering loss. Since joint angles directly affect the tool's articulation, this step ensures accurate motion reconstruction. The joint angles are differentiably optimized using gradient descent, subject to physical constraints. The update rule is:  
\begin{equation}
\mathbf{q}_{\text{updated}} = \mathbf{q}_{\text{current}} - \gamma \cdot \nabla_{\mathbf{q}} L_{\text{combined}}
\end{equation}

where:  
\begin{itemize}
    \item \( \mathbf{q}_{\text{updated}} \) is the new set of joint angles after optimization.
    \item \( \mathbf{q}_{\text{current}} \) is the current joint configuration.
    \item \( \nabla_{\mathbf{q}} L_{\text{combined}} \) is the gradient of the loss function with respect to the joint angles.
    \item \( \gamma \) is the learning rate for joint angle optimization, set to \(10^{-3}\).
\end{itemize}

To prevent infeasible joint configurations, we enforce joint limits using a constraint function:

\begin{equation}
\mathbf{q}_{\text{updated}} = \text{clamp}(\mathbf{q}_{\text{updated}}, \mathbf{q}_{\text{min}}, \mathbf{q}_{\text{max}})
\end{equation}

where \( \mathbf{q}_{\text{min}} \) and \( \mathbf{q}_{\text{max}} \) define the allowable joint range. This ensures the estimated joint angles remain within physically valid limits.

\section{Experiments}
\subsection{Experimental Setup}
The experimental setup is shown in Fig.~\ref{fig:overview}b, where a \gls{psm} of \gls{dvrksi} provides six \gls{dof} for motion. The \gls{dvrksi} system is developed based on the da Vinci Si robot, which is outfitted with a control system specifically designed for research purposes~\cite{kazanzides2014open}. In this study, a large needle driver is employed as the end-effector. Videos are captured using the \gls{ecm} of the system, which remains stationary during the procedure to provide a fixed camera frame. Additionally, we mount a wrist camera on the needle driver to provide additional visual context when training the imitation learning policy. Incorporating a wrist-mounted camera has been shown to improve policy performance by providing a more detailed local view of the manipulation task~\cite{kim2024surgical}. A pink suture pad is used as the background, while a white 2D suture pad and a surgical needle are positioned on top to conduct tissue-lifting and needle pick-up tasks.

\begin{figure*}[t]
    \begin{center}
    \vbox{ 
        \includegraphics[width=\textwidth]{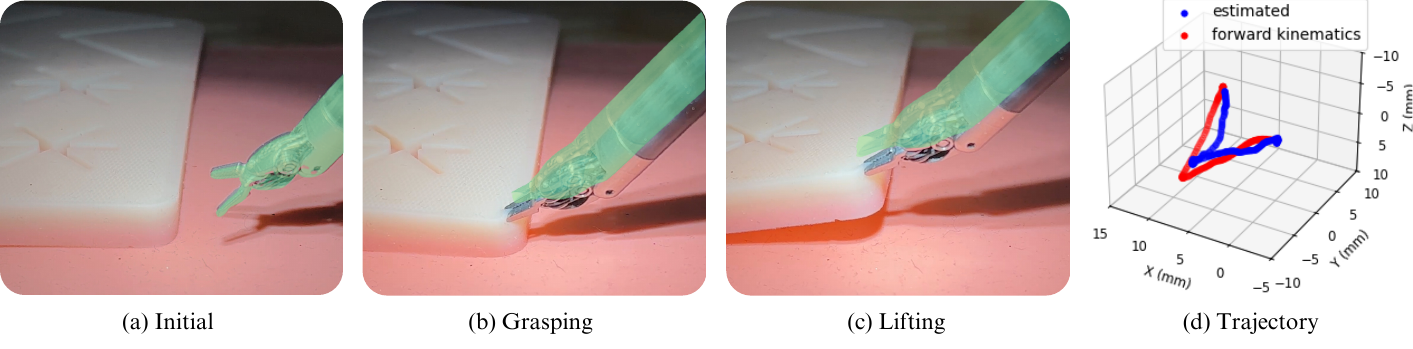}
    }
    \caption{The first three images show snapshots of the robot executing the estimated trajectory. A green mask overlay represents the corresponding tool pose from the original recorded video. (a) The robot starts from its initial pose. (b) The robot reaches and grasps the tissue. (c) The robot successfully lifts the tissue. (d) The end-effector trajectory estimated by our pipeline is compared to the ground truth trajectory obtained using forward kinematics.}
    \label{fig:tissue-lift-results}
    \end{center}
\end{figure*}

\subsection{Trajectory Extraction and Replay}
%% Experiment overview (datasets description, Experiment settings)
We first evaluate the feasibility of our pipeline by testing its ability to extract kinematic trajectories from monocular endoscopic videos. In this experiment, we recorded a video of a tissue-lifting task. The video is then processed by SurgiPose to extract actions. To ensure consistency between the recorded demonstration and the robot replay, the initial end-effector pose is stored and used to initialize the robot prior to execution. The extracted trajectory is then replayed on the \gls{dvrksi} to assess whether the vision-based kinematic estimation is sufficient to replicate the tissue-lifting task. We qualitatively evaluate the task execution by observing key stages of the motion (initial position, grasping, and tissue lifting) and quantitatively compare the extracted trajectory with ground truth trajectories derived via forward kinematics.

\subsection{Imitation Learning Policy Training and Evaluation}
After confirming the feasibility of using SurgiPose for trajectory extraction, we collected 220 demonstrations of the tissue-lifting task and 224 demonstrations of the needle pickup task, capturing synchronized video and ground-truth kinematics. Using our pipeline, we estimated kinematic trajectories solely from video data. These inferred trajectories, along with the corresponding video frames, were used to train an imitation learning policy. To evaluate the effectiveness of our approach, we compared the performance of policies trained with inferred kinematics against those trained with recorded ground-truth kinematics. Note that wrist camera images were also collected during data collection, but they were only used for training the imitation learning policy, not for inferring kinematic information.

Evaluation metrics for policy performance included task success rate and execution time. Preliminary results indicate that policies trained with video-inferred trajectories perform comparably to those trained with ground-truth kinematics, supporting the feasibility of our framework for vision-based surgical robot learning.

In addition to evaluating policy performance, we computed kinematic metrics to directly compare the estimated trajectories with ground-truth kinematics across all collected demonstrations. Specifically, we analyzed \textbf{Average Displacement Error (ADE)}, \textbf{Final Displacement Error (FDE)}, and \textbf{mean error in Cartesian coordinates (\(x, y, z\))}  to quantify the accuracy of our kinematic estimation pipeline.  

The Average Displacement Error (ADE) measures the mean Euclidean distance between the estimated trajectory and the ground-truth trajectory over all time steps:

\begin{equation}
\text{ADE} = \frac{1}{T} \sum_{t=1}^{T} \|\hat{\mathbf{p}}_t - \mathbf{p}_t\|
\end{equation}

where \( T \) is the total number of time steps, \( \hat{\mathbf{p}}_t \) is the estimated end-effector position at time step \( t \), and \( \mathbf{p}_t \) is the ground-truth position at time step \( t \).

The Final Displacement Error (FDE) quantifies the Euclidean distance between the estimated and ground-truth positions at the final time step:

\begin{equation}
\text{FDE} = \|\hat{\mathbf{p}}_T - \mathbf{p}_T\|
\end{equation}

where \( \hat{\mathbf{p}}_T \) and \( \mathbf{p}_T \) represent the estimated and ground-truth positions at the final time step \( T \), respectively.

The mean error in Cartesian coordinates is computed as the average per-axis difference between estimated and ground-truth positions:

\begin{equation}
\text{Mean Error} (x, y, z) = \frac{1}{T} \sum_{t=1}^{T} (\hat{\mathbf{p}}_t - \mathbf{p}_t)
\end{equation}

where the result is a vector representing the average error along each axis.
These metrics provide a comprehensive evaluation of our method’s accuracy in estimating tool trajectories from monocular video.

\noindent
\textbf{Policy Training:}
To train our imitation learning policies, we adopted action chunking with transformers (ACT)~\cite{zhao2023learning}, along with the hybrid-relative action representation proposed in SRT~\cite{kim2024surgical}. This representation encodes delta translations relative to the endoscope frame and delta rotations relative to the current end-effector frame, mitigating inaccuracies in the da Vinci robot’s joint angle measurements that could otherwise hinder policy learning.

For visual feature extraction, we employed a pre-trained EfficientNet-B3~\cite{tan2019efficientnet} as the image encoder. Training was conducted on an RTX 4090 GPU for approximately 20 hours.

\noindent
\textbf{Evaluation Setup:}
Each policy was evaluated over 10 trials of the tissue-lifting task. We measured the success rate, defined as the percentage of trials in which the robot successfully completed the task, and the completion time, which is the average time taken to complete the task. If a trial failed to complete the task, it was excluded from the calculation of the average task completion time.

\subsection{Evaluation on SurgRIPE and Ex Vivo Datasets}
To assess the applicability and generalizability of SurgiPose, we evaluate it on two datasets: the publicly available SurgRIPE dataset~\cite{xu2025surgripe} and a self-collected ex vivo cholecystectomy dataset.

The SurgRIPE dataset provides ground-truth absolute tool pose obtained using a keydot marker, which is later removed from images using a deep-learning inpainting model. While this dataset is primarily designed for benchmarking absolute pose estimation frameworks, it lacks ground-truth joint information, limiting its suitability for evaluating our method’s ability to estimate joint angles. Nonetheless, we use this dataset to qualitatively assess our model’s ability to infer 6-DoF tool pose from monocular images.

To further demonstrate the generalizability of our approach, we apply our pose estimation pipeline to a self-collected ex vivo cholecystectomy dataset. Unlike SurgRIPE, this dataset contains kinematic information, allowing us to evaluate our method on a different tool model. Specifically, we assess its performance on the ProGrasp forceps, which differs in both shape and articulation from the tools in SurgRIPE. This experiment aims to verify whether our method can adapt to different surgical tool geometries and motion patterns in real-world surgical settings.

\section{Results}
\subsection{Trajectory Replay Results}
Qualitative results of the trajectory replay experiment are shown in Fig. \ref{fig:tissue-lift-results}. The first three images in Fig. \ref{fig:tissue-lift-results} show snapshots of the robot following the estimated trajectory, with a green mask overlay representing the corresponding tool pose from the original recorded video. Visually, the executed trajectory closely aligns with the recorded motion, demonstrating that our method can extract meaningful kinematic information from video. However, minor deviations were observed, particularly in depth estimation, which occasionally caused the tool to move slightly farther from the camera than in the original execution. The trajectory plot on the right quantitatively compares the estimated end-effector trajectory to the ground truth obtained from forward kinematics. While the estimated trajectory follows a similar trend to the ground truth, slight discrepancies suggest that improvements in depth accuracy could further enhance the precision of the inferred kinematics.

\begin{figure*}[t]
    \begin{center}
    \vbox{ 
        \includegraphics[width=\textwidth]{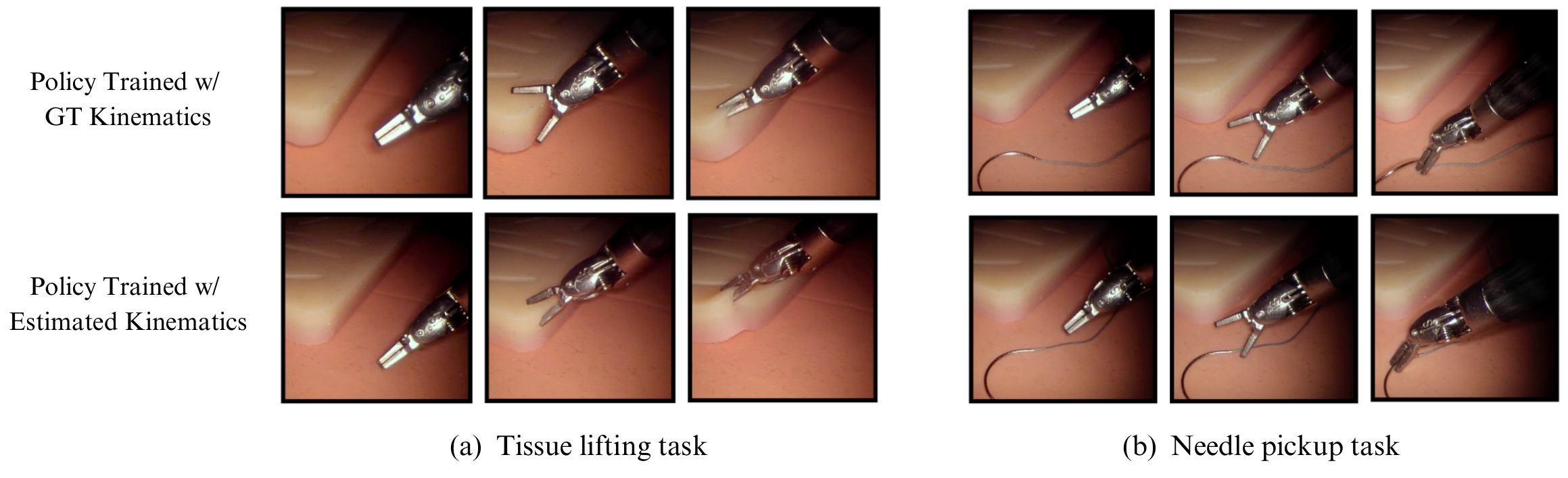}
    }
    \caption{Qualitative results for imitation learning experiment: Snapshots of key moments in the tissue-lifting experiment comparing policies trained with ground-truth and estimated kinematics. The top row shows the policy trained with ground-truth kinematics at three key moments: (1) initial pose, (2) grasping the tissue, and (3) lifting the tissue. The bottom row presents the same key moments for the policy trained with estimated kinematics directly from video.}
    \label{fig:IL-results}
    \end{center}
\end{figure*}

\subsection{Imitation Learning Results} The results in Table~\ref{tab:trajectory_comparison} provide insights into the accuracy of our differentiable rendering pipeline in estimating kinematic trajectories for the tissue lifting and needle pickup tasks. The average displacement error is 9.7 mm for tissue lifting and 12.0 mm for needle pickup, indicating that on average, the estimated trajectories deviate by these amounts from the ground truth. The final displacement error is slightly higher for tissue lifting (15.3 mm) compared to needle pickup (14.4 mm), suggesting that for tissue lifting, trajectory deviation tends to accumulate more over time, likely due to depth estimation errors.

\begin{table}[h]
    \centering
    \caption{Trajectory Estimation Errors for Tissue Lifting and Needle Pickup (mm)}
    \label{tab:trajectory_comparison}
    \begin{tabular}{lcc}
        \toprule
        Metric & Tissue Lifting (mm) & Needle Pickup (mm) \\
        \midrule
        Average ADE  & 9.7 $\pm$ 2.8 & 12.0 $\pm$ 2.8 \\
        Average FDE  & 15.3 $\pm$ 4.7 & 14.4 $\pm$ 4.5 \\
        Mean Error (x, y, z) & [-4.64, 0.25, 6.64] & [-6.84, 4.20, 5.61] \\
        Std Error (x, y, z) & [1.27, 1.40, 3.85] & [2.38, 1.48, 5.84] \\
        \bottomrule
    \end{tabular}
\end{table}

Analyzing the mean error in Cartesian coordinates reveals that the largest deviation occurs in the z-direction (depth dimension) for both tasks, with an average error of 6.6 mm for tissue lifting and 5.6 mm for needle pickup. Depth estimation remains particularly challenging as our method relies solely on monocular video, making it difficult to accurately infer scale and perspective. One possible source of error is that the differentiable rendering process may favor solutions that reduce visual discrepancy by making the surgical tool appear smaller in the image, as a smaller tool in the image can better match the observed image features, leading to a bias in predicting a tool trajectory further from the camera. The x- and y-direction errors are smaller in magnitude but still noticeable, with higher variability in the needle pickup task, as indicated by its larger standard deviations. This increased variability could be attributed to the more complex tool interactions required for needle manipulation compared to simple tissue lifting.

Despite these errors, the overall trajectory deviations remain within an acceptable range for policy training, as demonstrated by the comparable success rates between policies trained with estimated and ground-truth kinematics. Future work could focus on improving depth estimation by incorporating temporal consistency constraints or leveraging learned priors from large-scale surgical video datasets to refine trajectory predictions.

Fig.~\ref{fig:IL-results} shows qualitative results for our imitation learning experiment. Table \ref{tab:policy_comparison} shows the results of our experiment comparing the performance of two policies trained for tissue-lifting task and needle pickup task—one using ground-truth kinematics and the other using estimated kinematics.

\begin{table}[h]
    \centering
    \caption{Comparison of Policies Trained with Estimated and Ground-Truth Kinematics Across Different Tasks}
    \label{tab:policy_comparison}
    \begin{tabular}{lcccc}
        \toprule
        \multirow{2}{*}{Policy} & \multicolumn{2}{c}{Tissue Lifting} & \multicolumn{2}{c}{Needle Pickup} \\
        \cmidrule(lr){2-3} \cmidrule(lr){4-5}
        & Success & Time (s) & Success & Time (s) \\
        \midrule
        G.T. & 10/10 & 25.0 $\pm$ 3.9 & 8/10 & 23.5 $\pm$ 5.0  \\
        Est. (ours) & \textbf{7/10} & 33.0 $\pm$ 2.4 & \textbf{6/10} & 22.8 $\pm$ 2.9 \\
        \bottomrule
    \end{tabular}
\end{table}

For the tissue lifting task, the policy trained with ground-truth kinematics achieved a 100\% success rate, while the policy trained with estimated kinematics achieved a 70\% success rate. Although the estimated kinematics policy can successfully grasp the tissue in most cases, we observed failure cases where the robot tended to push the tissue away instead of lifting it, which aligns with our observation that depth estimation in our pipeline is less accurate.

In addition, we observed that the policy trained with estimated kinematics required more time to complete the task, taking an average of 33.0 seconds compared to 25.0 seconds for the ground-truth policy. This increased execution time suggests that inaccuracies in the estimated kinematics may lead to suboptimal motion strategies, requiring additional adjustments during execution.

For the needle pickup task, the policy trained with ground-truth kinematics achieved an 80\% success rate, while the policy trained with estimated kinematics reached 60\%, which is approximately 70\% of the baseline performance. During execution, we observed that both policies exhibited visual-servoing behavior, attempting to align the needle to the center of the opened jaws in the wrist camera view. This further underscores the importance of wrist camera input for learning dexterous manipulation tasks, as previously suggested in~\cite{kim2024surgical}. One notable limitation of the baseline policy was its inability to achieve a 100\% success rate, which could be attributed to the nature of the collected demonstrations. The dataset consisted primarily of perfect demonstrations, where the needle was picked up without errors, without including examples of recovery strategies. As a result, the policy lacks robustness against deviations that occur due to compounded execution errors. Expanding the dataset to include recovery strategies could enhance generalization and improve policy reliability under varying conditions.

Despite the observed performance gap, the comparable success rates between the two policies suggest that our approach provides a feasible alternative for training imitation learning policies without direct kinematic supervision. By extracting kinematics from monocular videos, we enable scalable imitation learning from existing surgical video datasets, reducing reliance on manually recorded kinematic trajectories. 

\subsection{Estimation Results on SurgRIPE and Ex Vivo Datasets}
\begin{figure}[t] 
\centering
\vbox{
\includegraphics[width=0.45\textwidth]{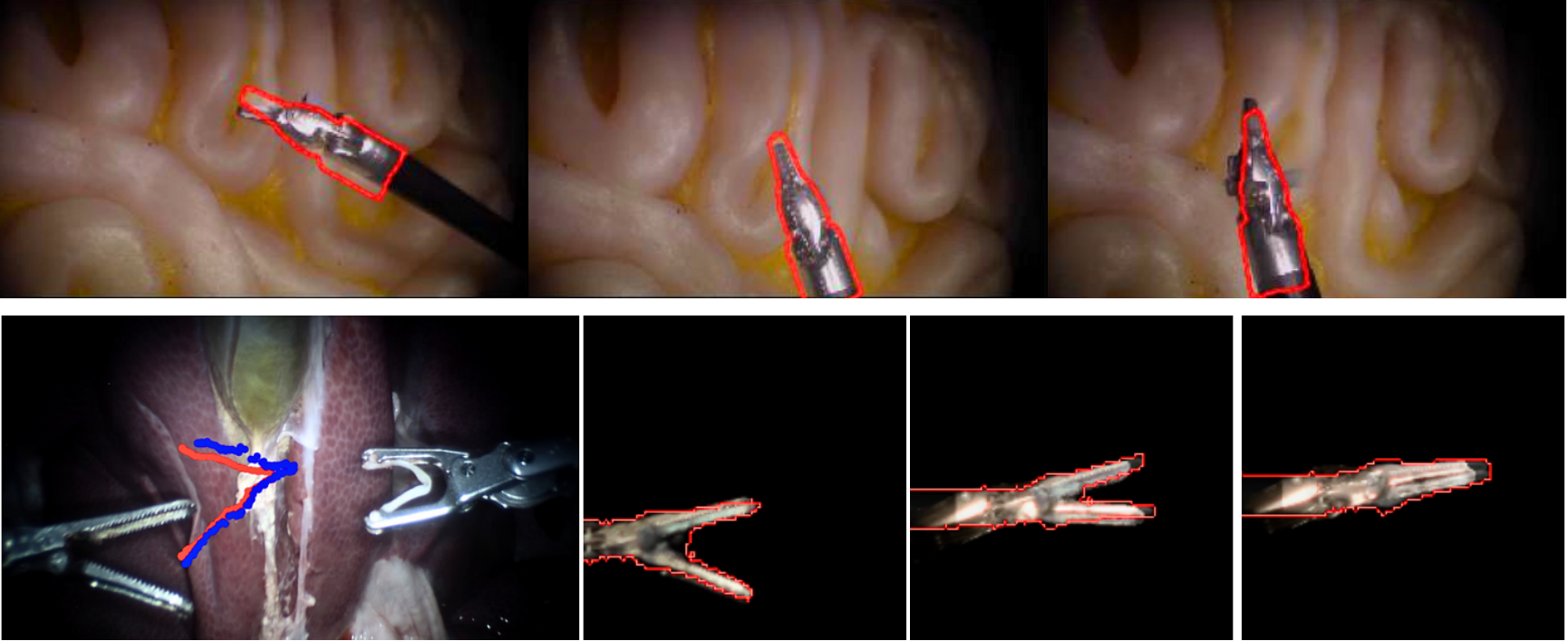} 
}
\caption{Qualitative results of our method applied to the SurgRIPE dataset~\cite{xu2025surgripe} and our self-collected dataset. The top row shows our estimated tool pose (silhouette) overlaid on images from the SurgRIPE dataset. The bottom row shows the results on the ex vivo dataset: the leftmost image compares our estimated trajectory (blue) with the ground truth trajectory projected onto the image (red), while the remaining three images shows the segmented tool with our estimated pose overlaid on top of it.} 
\label{fig:qualitative_results}
\end{figure}

Figure \ref{fig:qualitative_results} shows the qualitative results of our method applied to the dataset from~\cite{xu2025surgripe}. We overlay our estimated tool silhouette on random examples from the original dataset.  Due to the absence of ground truth joint information in the dataset, only a qualitative assessment is possible. We observe that our estimates correspond closely with the tool's pose and joint configuration. This preliminary experiment highlights the applicability of our method to other data, demonstrating its potential for generalization to different surgical environments.

We further validate our method on our self-collected ex vivo cholecystectomy dataset. Figure~\ref{fig:qualitative_results} shows a comparative visualization where we project the estimated trajectory (blue) and the ground-truth kinematics trajectory (red) onto the image. This projection provides an intuitive assessment of our model’s accuracy in recovering tool motion. Additionally, we present segmented frames from the video, where the estimated tool pose is overlaid as a red silhouette to highlight alignment with the observed tool motion. These results suggest that our method can generalize beyond a single dataset and tool type, demonstrating its adaptability to real surgical video data.

\section{Discussion and Conclusion}

Our results demonstrate that SurgiPose can reconstruct kinematic trajectories with an average displacement error of 9.7 mm for tissue lifting task and 12.0 mm for needle pickup task. Using the estimated kinematics, we trained imitation learning policies to perform the same tasks, achieving a 70\% success rate in tissue lifting and 60\% success rate in needle pickup. These results highlight the potential of using estimated kinematics for robot learning, showing that policies trained with inferred kinematics can achieve performance comparable to those trained with ground-truth kinematics.

However, there are some limitations and failure cases worth noting. First, depth estimation inaccuracies in our pipeline cause the policy trained with estimated kinematics to move further away from the camera when executing actions. Additionally, our method relies on continuous tool visibility, meaning any occlusion during the video can lead to errors in the estimated joint angles and trajectory. \revise{Moreover, our method assumes both jaws are visible in the first frame; occlusion or rotation may degrade initialization. This could be mitigated with symmetry priors or learning from partial inputs. The method also relies on accurate first-frame segmentation, though it is robust to minor noise due to grid-based initialization and optimization. Future work may improve robustness with temporal smoothing or learned segmentation models.} \deleted{Finally, }Lighting conditions can \revise{also} \deleted{significantly }affect estimation accuracy, as poor lighting reduces the clarity of visual cues necessary for precise depth and pose estimation. \revise{Finally, while we demonstrated generalizability on ex vivo cholecystectomy data, testing on textured phantoms with varied tissue properties would further validate robustness in diverse settings.}

Despite these challenges, this work demonstrates the feasibility of using a differentiable rendering pipeline to estimate kinematic information from surgical videos.
By enabling kinematic estimation from monocular surgical videos, our approach provides a scalable alternative to direct kinematic data collection, paving the way for large-scale demonstration learning and the development of autonomous surgical systems.

%%%%%%%%%%%%%%%%%%%%%%%%%%%%%%%%%%%%%%%%%%%%%%%%%%%%%%%%%%%%%%%%%%%%%%%%%%%%%%%%

%%%%%%%%%%%%%%%%%%%%%%%%%%%%%%%%%%%%%%%%%%%%%%%%%%%%%%%%%%%%%%%%%%%%%%%%%%%%%%%%


\section*{ACKNOWLEDGMENT}
Research reported in this paper was supported by NSF/FRR 2144348, NIH R56EB033807, and ARPA-H 75N91023C00048.

%%%%%%%%%%%%%%%%%%%%%%%%%%%%%%%%%%%%%%%%%%%%%%%%%%%%%%%%%%%%%%%%%%%%%%%%%%%%%%%%

\bibliographystyle{IEEEtran}
\bibliography{references}

\addtolength{\textheight}{-12cm}   % This command serves to balance the column lengths

\end{document}